\documentclass{article}
\usepackage{iclr2027_conference,times}

\usepackage{hyperref}
\usepackage{url}

\usepackage{placeins}
\usepackage{booktabs}
\usepackage{array}
\usepackage{multirow}
\usepackage{graphicx}

\usepackage{tikz}
\usepackage{xcolor}
\usepackage{amsmath}
\usepackage{amssymb}

\iclrfinalcopy

\usepackage{etoolbox}
\makeatletter
\patchcmd{\@maketitle}
  {\lhead{Published as a conference paper at ICLR 2027}}
  {\lhead{}}
  {}{}
\makeatother

\definecolor{unst}{RGB}{204,102,92}
\definecolor{stab}{RGB}{88,152,120}
\definecolor{dead}{RGB}{135,135,140}
\definecolor{cline}{RGB}{70,70,70}
\definecolor{axis}{RGB}{110,110,110}
\definecolor{thrline}{RGB}{60,60,60}
\definecolor{band}{RGB}{95,120,175}

\title{Measuring the Stability Assumption Behind Action Chunking}

\author{
  Aryan Goyal\\
  Independent Researcher\\
  \texttt{garyan18@gmail.com}
}

\begin{document}
\maketitle

\begin{abstract}
Action chunking improves the performance of policies learned by behavioural cloning, and several mechanisms have been proposed to explain why, including temporal consistency, horizon reduction, representation learning, and reduced error compounding. Every one concerns
the error the policy introduces, whether by reducing how much of it enters or
how often it re-enters the policy's own input. We instead study what happens to
an action error once it enters the system. At each state, we inject a small action error and measure how fast it grows or shrinks under two execution regimes: open-loop, where the rest of the chunk is replayed without replanning, and closed-loop, where the policy replans after the perturbation. The fitted rate labels each state as contracting, expanding, or unresolved. Across twelve manipulation tasks from three benchmark
suites, we find that confidently stable states are rare, while error
amplification is common among states whose propagation rate can be resolved.
We further find that the measured propagation rate depends strongly on the fitting horizon: amplification is typically front-loaded, so short windows can substantially overestimate longer-horizon propagation. Finally, we train predictors on these labels and find that a state's open-loop regime can be recovered from camera frames and proprioception alone, without a simulator, while its closed-loop propagation is only partially recoverable because it also depends on how the policy acts after the perturbation. These results suggest that error-compounding arguments alone do not provide a complete account of action chunking: neither passive open-loop dynamics nor policy replanning consistently contracts an injected error, and replanning rarely turns open-loop amplification into confident contraction. This also suggests that closed-loop reactivity should be trained explicitly, using perturbation- and tree-coverage-oriented training to expose the policy to deviations it must recover from, rather than expected to emerge reliably from standard imitation learning.

\end{abstract}




\section{Introduction}
\label{sec:intro}

Imitation-learned manipulation policies increasingly predict chunks of actions
rather than single steps \citep{zhao2023act, chi2023diffusion}. A chunked
executor plays $k$ actions open-loop and then replans, and $k$ is chosen once
per task. Chunked execution is widely reported to outperform single-step
execution, but the explanation is contested, and the accounts on offer attribute
the gain to different mechanisms.

Several mechanisms have been proposed to explain the advantage.
\emph{Temporal consistency}: modelling actions jointly matches correlated
demonstrator behaviour that independent sampling does not \citep{zhao2023act,
chi2023diffusion, li2025rlchunking}. \emph{Horizon reduction}: suboptimality
scales with the number of decisions \citep{ross2010efficient,
foster2024behavior}, and a chunk makes $k$ times fewer \citep{zhao2023act}.
\emph{Representation learning}: the harder training target yields a better
internal representation \citep{chi2023diffusion, torne2025longcontext}.
\emph{Non-Markovian expressivity}: action chunking can capture history-dependent demonstrator behaviour that a Markovian policy cannot \citep{lazzati2026chunking}. \emph{Delayed
conditioning}: predicting later actions from earlier observations can reduce the effect of the policy's own accumulated errors, an effect that can often be reproduced by delayed policies \citep{lazzati2026chunking}.
\emph{Implicit ensembling}: a chunked policy learns multiple temporal relationships between observations and actions, producing ensemble-like robustness \citep{lazzati2026chunking}.
\emph{Limited context}: chunking substitutes for history the policy cannot
condition on, and the advantage recedes once the context window is extended
\citep{zeng2026revisiting}.

Every one of these accounts concerns the error that enters the loop: how often
the policy introduces one, or how large it is when introduced. None of them
addresses what happens to the error afterwards, and that is arguably the more
consequential question, because small deviations are inevitable in learned
policies. The introduced error can be amplified, absorbed, or carried forward along the horizon of the task.

The theoretical bounds do speak to this, but they assume a single global
contraction rate for the dynamics rather than a rate that varies with state.
\citet{zhang2025chunking} assume that the true dynamics are open-loop
exponentially incrementally input-to-state stable (EISS) with a fixed decay
factor $\rho\in(0,1)$: roughly, $\rho$ determines how quickly a deviation
between trajectories decays from one step to the next. Under this assumption,
and with regularity controlled by the EISS constant $C_{\mathrm{ISS}}$, they
show that sufficiently long action chunks restore exponential stability of
the learned closed loop. The resulting chunk-length threshold depends on
$\rho$, with weaker contraction (larger $\rho$) requiring longer chunks.
In practice, however, the propagation of an error can depend strongly on the
state and task geometry, so a single global $\rho$ may not capture the local
error-propagation behaviour of a robot throughout an episode.

\citet{lazzati2026chunking} likewise derive their bound under deterministic
Lipschitz dynamics. In both analyses, the relevant dynamical constants are
assumed rather than measured and treated as global, whereas the propagation of
an action error may depend strongly on the state and task geometry. A manipulator
in free space and the same manipulator during contact, for example, need not
propagate an error at the same rate.

Replanning is the other way an error can be dealt with once it has entered,
which raises the same question about the deployed policy rather than the plant.
Closed-loop stability is a property of a controller paired with a plant, and the
pair that matters here is not the one the theory assumes. The expert paired with
the true dynamics, $(\pi^{\star}, f)$, is stable by construction, since a human
demonstrator sees the error and corrects it. What determines whether replanning
helps at deployment is the learned policy paired with the same dynamics,
$(\hat{\pi}, f)$, and \citet{zhang2025chunking} show that stability of the
former does not imply stability of the latter at single-step execution.
Behaviour cloning fits the expert's actions at the expert's states and
constrains little about what the policy does once it has drifted, which is
precisely where correction is required. If $(\hat{\pi}, f)$ is stable at a given
state then replanning there does contract the error and the argument holds; the
point is that this is an empirical property of the trained policy in its
environment, to be measured rather than inherited from the demonstrator.

Existing methods that vary the execution horizon at test time use signals for deciding
when to replan, rather than measuring what happens to an action error after it
enters the system. They trigger on model confidence
\citep{chopra2025everydayvla}, on attention weights
\citep{wang2026autohorizon}, on the entropy of the predicted actions
\citep{liang2026aac}, on the variance of the denoising process
\citep{dvac2026}, or on the agreement between chunks sampled at different
horizons \citep{jing2025moh}. Each of those is a proxy for how certain the
policy is about its own output rather than a measurement of how the plant treats
an action error, and none produces per-state labels that exist independently of
the controller consuming them.

\paragraph{Contributions.} We label, per state, how an injected action error
propagates over the steps that follow, on two axes: what the plant does with it
while the policy executes a chunk blind, and what the deployed policy does with
the same error when it is allowed to replan. The two correspond to the decision
the executor makes at that moment, since a longer chunk leaves the deviation to
the dynamics and a shorter one hands it to the policy, so the labels say which
of the two is doing the work at each state. This lets us ask whether the
assumption the chunking guarantees rest on, that the plant absorbs a small error
before the next replan, actually holds along the states these policies
encounter. We report the labels across three benchmark families, robomimic,
MimicGen and RoboCasa, spanning twelve tasks and two policy classes, at
demonstration states, at states the policy visits, and on the chunks the policy
itself plans. Finally we ask whether the regime is recoverable from what a robot can observe,
training predictors on camera frames, proprioception, and both together with the
actions about to be executed, and find that one axis is predictable while the
other is only partially recoverable


\section{Related Work}
\label{sec:related}

Recent work has proposed several explanations for why action chunking improves
imitation learning, including non-Markovian expressivity and delayed
conditioning \citep{lazzati2026chunking}, and the role of limited observation
context \citep{zeng2026revisiting}. \citet{zhang2025chunking} provide a
complementary control-theoretic analysis, showing that under open-loop
exponentially incrementally input-to-state stable dynamics with a fixed
contraction factor, sufficiently long action chunks prevent exponential error
compounding and yield corresponding trajectory-error bounds. Our work
complements these explanations by directly measuring how an injected action
error propagates at individual states, both when a chunk is executed without
replanning and when the policy is allowed to replan.

\section{Methods}
\label{sec:method}

\subsection{Measuring stability per state}

\begin{figure}
    \centering
    \includegraphics[width=1\linewidth]{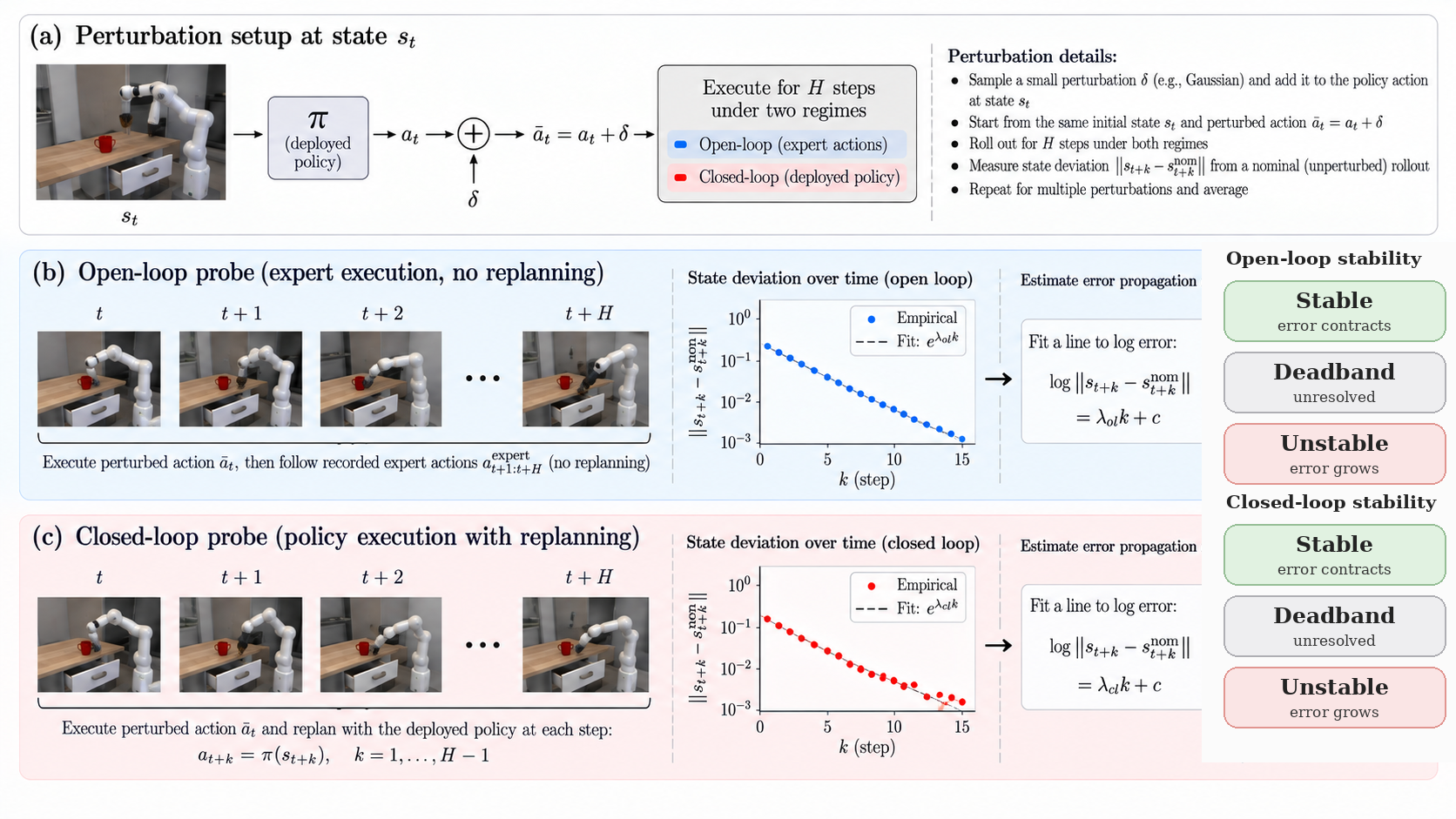}
 \caption {\textbf{Per-state measurement of open-loop and closed-loop error propagation.}
At a state $s_t$, a small action perturbation $\delta$ produces
$\tilde a_t=a_t+\delta$. We measure the resulting deviation from a nominal
rollout under two regimes: (a) open-loop, where the remainder of the recorded
action chunk is replayed without replanning, and (b) closed-loop, where the
deployed policy replans at each subsequent step. Nominal and perturbed
closed-loop branches share the same policy randomness. For each regime, we fit
the log state deviation to obtain a propagation rate $\lambda_{\mathrm{ol}}$ or
$\lambda_{\mathrm{cl}}$, which is classified as stable, deadband, or unstable
using the amplification and uncertainty gates.}

\end{figure}

Using the labeled states, we ask a simple question: starting from here, if the
executed action is slightly wrong, does the resulting deviation grow or shrink
over the steps that follow? The exponential form is motivated by the
incremental-stability bounds discussed above: for a small initial deviation,
such bounds describe propagation of the form
$|\Delta y_j| \propto \rho^j|\Delta y_0|$. Equivalently, since
$\rho^j=\exp(j\log\rho)$, the logarithm of the deviation should vary
approximately linearly with the number of steps. We therefore use a local
exponential model rather than assuming a global contraction factor, writing
$y_j$ for a task metric evaluated $j$ steps after the state along a nominal
execution and $y_j^{(n)}$ for the same quantity along the $n$-th perturbed
branch:
\begin{equation}
\log \lVert y_j - y_j^{(n)} \rVert
;\approx;
\alpha_n + \lambda_n j,
\label{eq:lambda}
\end{equation}
where $\alpha_n$ absorbs the initial magnitude of the perturbation and other
branch-specific scale factors, while $\lambda_n$ is the local
error-propagation rate. In the ideal exponential case, $\lambda_n$ plays the
role of $\log\rho$: a positive value indicates growth of the deviation, a
negative value indicates decay, and a value near zero indicates that the
deviation is approximately preserved. We repeat this fit across $N$ perturbed
branches and aggregate their slopes to obtain the mean rate and its standard
error. Estimating $\lambda_n$ requires a window over which to fit it, written
$K$, which is a property of the measurement rather than of the executed chunk.

\textbf{Open-loop stability.}
Since this is a property of the dynamics and task geometry rather than of a
policy, no trained policy is required to measure it. At a state $s_t$, we
replay the recorded actions without re-planning, perturbing only the
translation components of the first action with Gaussian noise of scale
$\sigma_u$. We set $\sigma_u$ to the action prediction error of the trained
policy on held-out demonstrations, providing a perturbation magnitude
representative of the policy's typical action error. The remaining actions are
replayed unchanged, isolating how an initial action error propagates through
the dynamics.

\textbf{Closed-loop stability.}
Closed-loop stability is a property of the plant together with the deployed
policy and therefore requires a trained policy. We evaluate it using the same
perturbation construction as in the open-loop setting, with one key difference:
the policy replans at every subsequent step after the initial perturbation.
Consequently, both the nominal and perturbed branches are policy-driven.
Replaying the demonstrator actions on either branch would confound the effect
of perturbation propagation with the discrepancy between the learned policy and
the demonstrator.

Because the policy may be stochastic, we use \emph{common random numbers} to
isolate the effect of the injected perturbation from variability due to policy
sampling. For each initial state, we generate $N$ paired nominal and perturbed
rollouts. Within each pair, the two branches use the same sequence of policy
random variables. The nominal branch executes the policy from the unperturbed
state, while the perturbed branch receives the initial action perturbation and
then replans from its perturbed state using the same policy random draws as its
nominal counterpart. The perturbation is applied only to the translation
components of the initial action.

For each paired branch, we measure the separation between the nominal and
perturbed trajectories over the fitting window and estimate the propagation rate
using the same log-distance fit described above. The closed-loop rate and its
standard error are obtained by averaging the branch-wise rates and using their
sample variation. This paired construction isolates the effect of the injected
perturbation without estimating or subtracting a separate stochastic-control
rate.

\textbf{State and action sources.}
We evaluate the open-loop probe at three stamp types, differing in the state
distribution and the source of the replayed actions. \emph{Demonstration} stamps
use recorded demonstration states and actions; \emph{rollout} stamps use states
reached by the policy and the actions it executed; and \emph{policy} stamps use
replan boundaries and the action chunk emitted by the policy. This gives two
controlled comparisons: demonstration versus rollout changes the state
distribution, while rollout versus policy changes the action source. Closed-loop
probes are evaluated at the same policy stamps, allowing the two execution
regimes to be compared at identical states. Table~\ref{tab:stamps} summarizes
the resulting stamp types and fitting windows.

\begin{table}[h]
\centering
\footnotesize
\setlength{\tabcolsep}{4pt}
\begin{tabular}{@{}l >{\raggedright\arraybackslash}p{2.3cm}
>{\raggedright\arraybackslash}p{3.2cm} c@{}}
\toprule
\textbf{Stamp type} & \textbf{State from} & \textbf{Actions used} & \textbf{$K$} \\
\midrule
demonstration & demonstration file &
demonstrator actions, as recorded & 24 \\[0.4em]

rollout & policy rollout &
policy actions, as recorded & 24 \\[0.4em]

policy chunk & policy rollout, at a replan boundary &
chunk emitted by the policy & 8 or 24 \\[0.4em]

closed loop & policy rollout, at a replan boundary &
policy replans at every step & 24 \\
\bottomrule
\end{tabular}
\caption{The four stamp types used in the analysis. $K$ denotes the fitting
window. For policy-chunk stamps, $K$ is capped by the emitted chunk length,
which is 8 actions for the diffusion policies and up to 24 on RoboCasa.}
\label{tab:stamps}
\end{table}

\textbf{Deciding a label.}
We apply the same two gates to both regimes. First, we ask whether the measured
rate is large enough to represent meaningful amplification or decay: this
requires $|\bar{\lambda}| \geq \delta_m$, where
$\delta_m=\ln(m)/K$ is the rate corresponding to an $m$-fold change over the
$K$-step measurement window. Second, we ask whether the sign of the measured
rate is sufficiently well resolved, requiring $|\bar{\lambda}| \geq
t\,\mathrm{SE}$. A stamp that passes both gates is labelled \emph{stable} when
$\bar{\lambda}<0$ and \emph{unstable} when $\bar{\lambda}>0$; stamps that fail
either gate are assigned to the \emph{deadband}.




\subsection{What each regime tells us}

The two stability measurements provide an empirical way to assess how chunk
execution contributes to error compounding. Open-loop stability tells us whether
an initial action error is amplified or attenuated while a chunk plays out,
while closed-loop stability tells us whether replanning suppresses that same
deviation. The resulting combinations therefore indicate, from the perspective
of error propagation alone, when continuing a chunk allows an error to grow and
when replanning can interrupt or reduce that growth. They are not, by
themselves, prescriptions for the globally preferred chunk length or task
performance.

\begin{table}[h]
\centering
\small
\begin{tabular}{p{2.4cm}|p{3.1cm}p{3.1cm}p{3.1cm}}
\toprule
& \textbf{CL stable} & \textbf{CL deadband} & \textbf{CL unstable} \\
\midrule
\textbf{OL stable} &
both modes are non-amplifying; little propagation-based reason to shorten &
no clear propagation-based preference &
replanning introduces amplification; continued execution may avoid it \\[0.5em]
\textbf{OL deadband} &
replanning provides evidence of contraction; shorter horizons may help &
neither regime is resolved &
replanning shows amplification; shortening should be weighed against other considerations \\[0.5em]
\textbf{OL unstable} &
replanning contracts an otherwise growing error; shortening has direct propagation-based motivation &
shortening removes the measured open-loop amplification but gives no evidence of contraction &
both regimes amplify the error; shortening alone does not resolve the measured propagation problem \\
\bottomrule
\end{tabular}
\caption{Interpretation of the joint open- and closed-loop stability labels.
The entries describe what the measured error-propagation behaviour implies for
the execution-horizon decision. They are not prescriptions for chunk length:
stability is one factor in that decision and does not determine task performance,
action quality, or the globally preferred horizon.}
\label{tab:regimes}
\end{table}

The bottom-left cell is the clearest case in which the stability measurements
provide actionable evidence: an error grows when the chunk is allowed to play
out, but contracts when the policy replans. In that setting, shortening the
execution horizon has a direct propagation-based rationale because it gives the
policy earlier opportunities to correct the deviation. More generally, the
table should be read as describing the error-propagation consequences of the two
execution regimes, rather than as determining the optimal chunk length on its own.

\subsection{Predicting the regime from observation}
\label{sec:predictors}

Since the open-loop rate is determined by the dynamics and task geometry
rather than by policy feedback, it is a property of the state the robot is in.
The policy used to reach that state does not affect how the plant propagates the
error once the executed actions are fixed. This makes it reasonable to ask
whether it is visible: does an observation of the current state, or a short
history of observations, carry enough evidence to say what the next several
steps will do with an injected error? We therefore ask whether $\lambda$ can be
recovered from what the robot can see, comparing two variants that differ only
in what they are given: perception alone, and perception together with the
actions about to be executed.

\textbf{Backbone.} Both variants use a frozen V-JEPA~2 ViT-L encoder
\citep{assran2025vjepa2} over a short window of camera frames ending at $t$,
together with proprioception over the same window. We use V-JEPA~2 because its
latent predictive objective is designed to capture spatiotemporal structure
relevant to understanding and predicting physical interactions, making it a
natural visual representation for our task of predicting how the robot's state
will evolve after an action perturbation. A learned query attends over
the visual tokens, while proprioception is encoded by a small recurrent
network; the resulting features are concatenated and passed to a two-layer
head. The encoder is never fine-tuned, so the variants differ only in their
inputs rather than backbone capacity. Dimensions, optimizer, and training
schedule are given in Appendix~\ref{app:predictor}.

\textbf{Outputs.}
Each head predicts both a continuous estimate of $\lambda$ and a logit for the
unstable class. The regression provides the measured rate, while the classifier
predicts the gated stability label. When both axes are labeled at the same
stamps, the same trunk produces separate outputs for each axis. We keep the
axes separate rather than predicting their conjunction, since the conjunction
classes are sparse.

\textbf{Variants.}
\emph{P} receives frames and proprioception only. It therefore tests whether the
regime is recoverable from the observed configuration. \emph{A} additionally
receives the action chunk under evaluation. This removes the mismatch between
the predictor input and the measurement, since $\lambda$ is defined for the
specific actions that are replayed. Thus, comparing P and A distinguishes
information available from the scene from information contained in the
planned actions.



\section{Experimental Results}
\label{sec:result}

\subsection{Datasets}
\label{sec:datasets}

We label three benchmark families, all built on \texttt{robosuite}
\citep{zhu2020robosuite} and MuJoCo \citep{todorov2012mujoco}, covering twelve
tasks and two policy classes, with both human and generated demonstrations.

\textbf{robomimic} \citep{robomimic2021}. Four proficient-human teleoperation
tasks: \texttt{lift}, \texttt{can}, \texttt{square}, and \texttt{tool\_hang},
spanning simple pick-and-place through high-precision insertion. Policies are
per-task diffusion policies \citep{chi2023diffusion}.

\textbf{MimicGen} \citep{mandlekar2023mimicgen}. Demonstrations generated by
transforming and replaying human source segments into new object poses,
retained by rejection sampling on success. We label \texttt{coffee},
\texttt{nut\_assembly}, \texttt{square\_d0}, and \texttt{stack\_d0} on the
demonstration axis. The first two additionally have trained diffusion policies
and therefore carry the rollout, policy, and closed-loop axes; the latter two
have no trained policy and are demonstration only. For these two tasks,
$\sigma_u$ is a proxy taken from a robomimic policy and is marked accordingly.

\textbf{RoboCasa} \citep{nasiriany2024robocasa}. Four kitchen tasks:
\texttt{OpenCabinet}, \texttt{OpenDrawer}, \texttt{PickPlaceCounterToCabinet},
and \texttt{TurnOnSinkFaucet}, each with 200 demonstrations. The policy is the
public GR00T~N1.5 checkpoint \citep{nvidia2025groot}, a multitask generalist
that we do not train. Demonstration, policy, and closed-loop axes are labeled.
There is no separate rollout corpus here, so the demonstration-to-policy
comparison on this family has no intermediate control.

\subsection{Three sources of open-loop states}
\label{sec:results-ol}
Across all three stamp types, unstable states are more common than stable states among measurements that clear the gates, while most stamps fall in the deadband. The deadband does not imply stability: it arises when the measured rate is either too small to exceed the amplification threshold or too uncertain to resolve its sign. Thus, open-loop execution does not generally avoid error growth through passive contraction. This also lets us test whether policies simply choose to commit at states that are naturally stable: if that were the case, policy-chunk stamps should contain more stable and fewer unstable states than the broader state distributions. 

Within \textsc{robomimic} demonstrations, the median propagation rate increases
from can (\(0.009\)) to lift (\(0.015\)), square (\(0.017\)), and tool hang
(\(0.026\)). This ordering is consistent with a trend toward more
contact-sensitive and precision-demanding manipulation: the largest rate occurs
on tool hang, where successful execution requires high-precision insertion.
Thus, states encountered in these more constrained manipulation regimes tend to
show stronger error amplification, although these medians do not establish
that task difficulty itself causes the increase. Rollout medians are close to
their demonstration counterparts (lift \(0.019\) vs.\ \(0.015\); tool hang
\(0.021\) vs.\ \(0.026\)), suggesting that moving from expert states to
policy-visited states does not substantially change how the plant propagates a
small error.

The \(G_2\) keep rate is \(90\text{--}99\%\) at \(K=24\). Thus, among stamps
that clear Gate 1, most also have a sufficiently resolved sign, so the large
deadband is not primarily a consequence of statistical uncertainty. Instead,
it is driven largely by rates that are too small to cross the amplification
threshold. For example, only 543 of 2,485 lift demonstration stamps clear Gate
1, meaning that \(78\%\) of stamps have
\(|\lambda|<0.0289\). This shows that most states do not exhibit confidently
measurable amplification or contraction at the chosen threshold, even though
the resolved subset is predominantly unstable.

\subsection{Closed-loop stability labels}
\label{sec:results-cl}

Closed-loop stability is substantially rarer than passive contraction under the same perturbation measurement. Across robomimic, MimicGen, and RoboCasa, the labels contain only a small number of confidently stable stamps, with most stamps assigned to the deadband or unstable classes. At \(m=2\), the stable class contains only 3--88 stamps per task, despite hundreds of non-stable stamps. 

In this sense, closed-loop execution can be worse than open-loop commitment: it trades away the temporal commitment of a chunk for reactivity without reliably reducing error propagation.

These results also expose a gap in standard imitation training. The task-specific diffusion policies are trained directly from expert demonstrations, without perturbation augmentation or explicit recovery supervision. Such training constrains the policy primarily on the demonstrated state distribution, but provides little supervision for recovery after a deviation. This suggests that closed-loop contraction may need to be encouraged explicitly by training on perturbed states and their corresponding recovery actions. This is consistent with prior work on noise and perturbation injection, including DART \citep{Laskey2017DART} and recent work by Simchowitz et al. \citep{simchowitz2025pitfalls}, where injecting perturbations during training exposes the policy to off-demonstration states and reduces compounding error. In particular, isotropic perturbations can expose the policy to deviations across action directions rather than only along the errors present in the demonstrations. Our results therefore suggest that closed-loop stability is unlikely to emerge reliably from imitation learning alone.


\begin{figure}[t]
\centering
\resizebox{\textwidth}{!}{\input{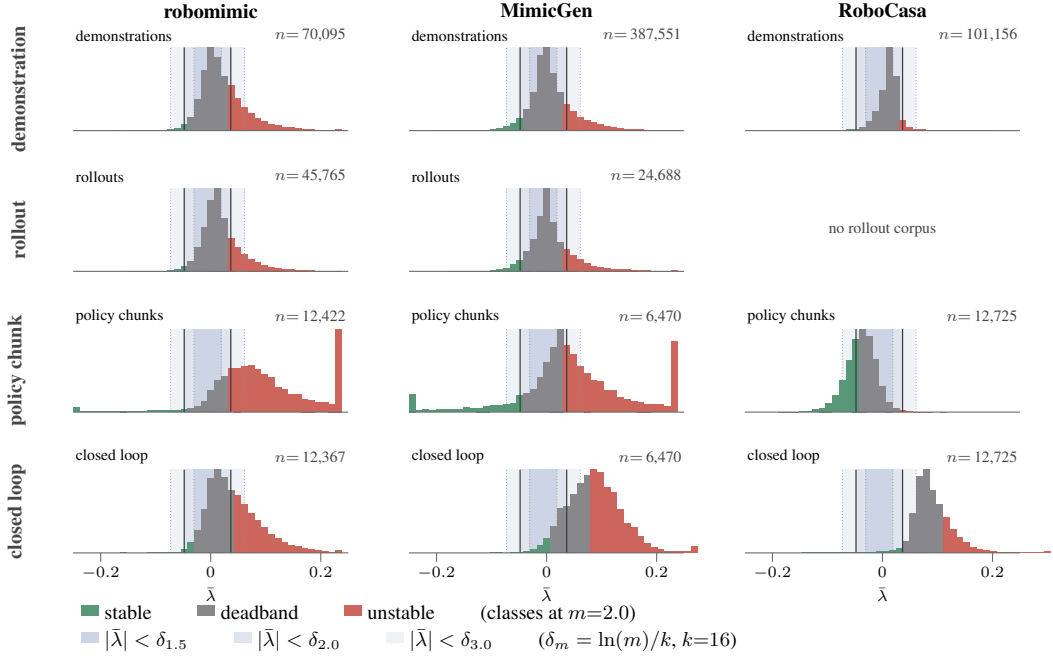}}
\caption{Distributions of the measured propagation rate $\lambda_{\mathrm{mean}}$,
pooled within each benchmark family. Columns correspond to robomimic,
MimicGen, and RoboCasa; rows correspond to demonstration, rollout,
policy-chunk, and closed-loop stamps. Stable, deadband, and unstable
regions are shown by the bar segments. Vertical lines show the reference
Gate-1 thresholds $\pm\ln(m)/16$ for $m\in\{1.5,2.0,3.0\}$; these are
shown as a common visual reference and do not replace the window-specific
threshold used for labeling. The demonstration, rollout, and closed-loop
rates are fitted over $K=24$, whereas the policy-chunk rate is capped by
the emitted chunk length.}
\label{fig:ol-hist}
\end{figure}

\label{sec:chunking-interpretation}

\begin{table}[t]
\centering\small\setlength{\tabcolsep}{4pt}
\begin{tabular}{llrrrrlrrr}
\toprule
Dataset & Stamps & $\sigma_u$ & $K$ & $n$ & median & p5--p95 &
$\lambda_{\mathrm{se}}$ & G1 pass & G2 keep \\
\midrule
\multicolumn{10}{l}{\emph{robomimic} — per-task diffusion policies} \\
lift & demos & 0.165 & 24 & 2{,}485 & .015 & $-$.012--.184 & .0076 & 543 & 96.7\% \\
lift & rollouts & 0.165 & 24 & 2{,}585 & .019 & $-$.020--.127 & .0084 & 690 & 97.0\% \\
lift & policy & 0.165 & 8 & 3{,}056 & .090 & .021--.458 & .0216 & 2{,}620 & 83.8\% \\
can & demos & 0.225 & 24 & 9{,}254 & .009 & $-$.032--.115 & .0059 & 1{,}844 & 90.2\% \\
can & rollouts & 0.225 & 24 & 8{,}360 & .019 & $-$.022--.116 & .0078 & 2{,}511 & 95.5\% \\
can & policy & 0.225 & 8 & 3{,}049 & .073 & $-$.016--.536 & .0172 & 2{,}169 & 87.0\% \\
square & demos & 0.220 & 24 & 12{,}723 & .017 & $-$.026--.128 & .0082 & 4{,}004 & 93.5\% \\
square & rollouts & 0.220 & 24 & 10{,}229 & .018 & $-$.029--.104 & .0067 & 2{,}145 & 92.5\% \\
square & policy & 0.220 & 8 & 3{,}116 & .091 & $-$.050--.385 & .0253 & 2{,}553 & 81.2\% \\
tool\_hang & demos & 0.164 & 24 & 45{,}633 & .026 & $-$.025--.114 & .0084 & 15{,}533 & 95.9\% \\
tool\_hang & rollouts & 0.164 & 24 & 24{,}591 & .021 & $-$.024--.101 & .0077 & 6{,}843 & 95.9\% \\
tool\_hang & policy & 0.164 & 8 & 3{,}201 & .113 & $-$.015--.348 & .0216 & 2{,}584 & 89.3\% \\
\midrule
\multicolumn{10}{l}{\emph{MimicGen} — per-task diffusion policies} \\
coffee & demos & 0.114 & 24 & 332{,}981 & .007 & $-$.047--.103 & .0072 & 83{,}698 & 95.3\% \\
coffee & rollouts & 0.114 & 24 & 12{,}384 & .009 & $-$.043--.105 & .0069 & 3{,}022 & 97.6\% \\
coffee & policy & 0.114 & 8 & 3{,}225 & .045 & $-$.134--.301 & .0191 & 1{,}990 & 82.1\% \\
nut\_assembly & demos & 0.103 & 24 & 33{,}251 & .009 & $-$.043--.097 & .0076 & 8{,}814 & 93.7\% \\
nut\_assembly & rollouts & 0.103 & 24 & 12{,}304 & .006 & $-$.046--.099 & .0078 & 3{,}047 & 94.2\% \\
nut\_assembly & policy & 0.103 & 8 & 3{,}245 & .046 & $-$.121--.329 & .0230 & 2{,}026 & 77.9\% \\
square\_d0 & demos & 0.220$^{\dagger}$ & 24 & 12{,}874 & .020 & $-$.024--.122 & .0083 & 4{,}291 & 94.2\% \\
stack\_d0 & demos & 0.203$^{\dagger}$ & 24 & 8{,}445 & .019 & $-$.014--.115 & .0069 & 1{,}595 & 96.9\% \\
\midrule
\multicolumn{10}{l}{\emph{RoboCasa} — GR00T N1.5} \\
OpenCabinet & demos & 0.565 & 24 & 33{,}238 & .016 & $-$.022--.038 & .0062 & 1{,}554 & 95.1\% \\
OpenCabinet & policy & 0.565 & 24 & 3{,}250 & .019 & $-$.040--.051 & .0060 & 360 & 96.7\% \\
OpenDrawer & demos & 0.426 & 24 & 24{,}226 & .018 & $-$.028--.042 & .0056 & 1{,}720 & 95.8\% \\
OpenDrawer & policy & 0.426 & 24 & 3{,}250 & .020 & $-$.043--.089 & .0069 & 614 & 95.8\% \\
PickPlaceToCab. & demos & 0.479 & 24 & 23{,}724 & .020 & $-$.006--.046 & .0048 & 1{,}420 & 99.4\% \\
PickPlaceToCab. & policy & 0.479 & 24 & 3{,}038 & .019 & $-$.035--.064 & .0054 & 430 & 99.3\% \\
TurnOnSinkFaucet & demos & 0.299 & 24 & 19{,}968 & .023 & $-$.013--.050 & .0047 & 1{,}553 & 98.5\% \\
TurnOnSinkFaucet & policy & 0.299 & 24 & 3{,}187 & .022 & $-$.021--.087 & .0061 & 483 & 96.7\% \\
\bottomrule
\end{tabular}
\caption{$\lambda_{\mathrm{mean}}$ on the open-loop axis, per dataset and stamp
type. $\sigma_u$ is the probe scale, the policy's measured action RMSE; $K$ is
the fitting window, bounded on policy stamps by the emitted chunk.
$\lambda_{\mathrm{se}}$ is the median per-stamp standard error over the eight
branches. \emph{G1 pass} counts stamps clearing Gate~1 at $m{=}2.0$ and
\emph{G2 keep} the fraction of those also clearing the $t$-test. IQR and the
remaining quantiles are in Table~\ref{tab:app-ol-quantiles}. Policies: per-task
diffusion policies on robomimic and MimicGen, GR00T~N1.5 on RoboCasa.
$^{\dagger}$ proxy $\sigma_u$; no policy was trained for these tasks, so they
have no rollout or policy stamps.}
\label{tab:ol-dist}
\end{table}


\subsubsection{Comparison at a matched fitting window}
\label{sec:ol-window-para}

\begin{table}[t]
\centering\small\setlength{\tabcolsep}{5pt}
\begin{tabular}{llrrrrrr}
\toprule
& & \multicolumn{3}{c}{$K{=}24$} & \multicolumn{3}{c}{$K{=}8$ (matched)} \\
\cmidrule(lr){3-5}\cmidrule(lr){6-8}
Dataset & Stamps & S & D & U & S & D & U \\
\midrule
\multicolumn{8}{l}{\emph{robomimic}} \\
lift & demos & 0.0 & 78.9 & 21.1 & 0.0 & 30.3 & 69.7 \\
lift & rollouts & 1.4 & 74.1 & 24.5 & 1.7 & 34.8 & 63.5 \\
lift & policy & -- & -- & -- & 0.2 & 28.1 & 71.7 \\
can & demos & 2.9 & 82.0 & 15.1 & 1.4 & 38.0 & 60.6 \\
can & rollouts & 1.7 & 71.3 & 27.0 & 1.6 & 34.4 & 64.0 \\
can & policy & -- & -- & -- & 1.7 & 38.1 & 60.2 \\
square & demos & 1.5 & 70.6 & 27.9 & 1.5 & 36.1 & 62.4 \\
square & rollouts & 2.0 & 80.6 & 17.4 & 2.0 & 29.3 & 68.7 \\
square & policy & -- & -- & -- & 2.4 & 33.5 & 64.1 \\
tool\_hang & demos & 1.4 & 67.4 & 31.3 & 1.2 & 33.2 & 65.6 \\
tool\_hang & rollouts & 1.4 & 73.3 & 25.3 & 1.1 & 31.2 & 67.7 \\
tool\_hang & policy & -- & -- & -- & 1.9 & 28.0 & 70.1 \\
\midrule
\multicolumn{8}{l}{\emph{MimicGen}} \\
coffee & demos & 5.5 & 76.1 & 18.4 & 5.2 & 52.8 & 42.0 \\
coffee & rollouts & 4.9 & 76.2 & 18.9 & 5.5 & 48.6 & 45.9 \\
coffee & policy & -- & -- & -- & 5.8 & 49.4 & 44.8 \\
nut\_assembly & demos & 4.7 & 75.2 & 20.2 & 6.2 & 53.7 & 40.1 \\
nut\_assembly & rollouts & 5.4 & 76.7 & 17.9 & 6.8 & 51.5 & 41.7 \\
nut\_assembly & policy & -- & -- & -- & 6.5 & 51.3 & 42.2 \\
square\_d0 & demos & 1.6 & 68.6 & 29.8 & 1.4 & 34.9 & 63.7 \\
stack\_d0 & demos & 0.3 & 81.7 & 18.0 & 0.4 & 25.1 & 74.5 \\
\midrule
\multicolumn{8}{l}{\emph{RoboCasa}} \\
OpenCabinet & demos & 1.2 & 95.6 & 3.2 & 2.3 & 23.3 & 74.4 \\
OpenCabinet & policy & 4.1 & 89.3 & 6.6 & 1.3 & 17.1 & 81.6 \\
OpenDrawer & demos & 2.5 & 93.2 & 4.3 & 3.4 & 19.3 & 77.3 \\
OpenDrawer & policy & 4.6 & 81.9 & 13.5 & 2.2 & 21.8 & 76.0 \\
PickPlaceToCab. & demos & 0.1 & 94.0 & 5.9 & 0.8 & 14.5 & 84.7 \\
PickPlaceToCab. & policy & 3.5 & 85.9 & 10.6 & 0.4 & 16.2 & 83.4 \\
TurnOnSinkFaucet & demos & 1.0 & 92.3 & 6.7 & 1.3 & 14.2 & 84.5 \\
TurnOnSinkFaucet & policy & 0.8 & 85.3 & 13.8 & 4.0 & 13.8 & 82.2 \\
\bottomrule
\end{tabular}
\caption{Open-loop label shares (\%) at $m{=}2.0$, $t{=}2.365$, at the stored
window and refit at $K{=}8$ on the same stamps. S, D, U = stable, deadband,
unstable. Policy stamps on robomimic and MimicGen are emitted as 8-action
chunks and have no $K{=}24$ entry. }
\label{tab:ol-shares-window}
\end{table}


The policy-chunk axis requires a separate treatment of the fitting window.
The diffusion policies on robomimic and MimicGen emit eight-action chunks, so
their open-loop divergence can only be fit over $K{=}8$. We use $K{=}24$ as the
primary window for demonstration and rollout stamps because it provides a
longer and more stable fit, but this makes the policy axis quantitatively
incomparable with the other two. Table~\ref{tab:ol-shares-window} therefore
refits the same stored per-branch log-distance matrices for demonstration and
rollout stamps at $K{=}8$. No additional simulation is required: the $K{=}8$
and $K{=}24$ columns contain the same stamps, evaluated over different portions
of the same trajectories. Thus, when comparing all three state sources on a
common basis, the matched $K{=}8$ results are the appropriate comparison.

Across datasets and state
sources, shortening the window from $K{=}24$ to $K{=}8$ moves a large fraction
of stamps from the deadband into the unstable class, while the stable fraction
changes little. For example, on robomimic lift demonstrations, the unstable
share increases from $21.1\%$ at $K{=}24$ to $69.7\%$ at $K{=}8$, while the
stable share remains $0.0\%$. The same qualitative shift occurs across the
other robomimic, MimicGen, and RoboCasa conditions. Thus, the conclusion that
open-loop contraction is uncommon is robust to the choice of window, but the
size of the unstable population is not.

This dependence arises because the divergence is typically front-loaded. The
perturbed branches separate rapidly during the first few steps and then tend
to flatten rather than continue growing at the same rate. A short window
therefore fits primarily the initial rise and produces a larger positive
$\lambda$, whereas a longer window averages that rise with the subsequent
flat region and pulls $\lambda$ toward zero. Because the later trajectory does
not generally reverse the initial separation, the additional steps move
stamps predominantly into the deadband rather than into the stable class.
Accordingly, $\lambda$ should be interpreted as a summary of error propagation
over the specified fitting interval, rather than as a time-invariant rate
governing the entire trajectory.

This distinction is important when relating the measurement to chunk length.
At $K{=}8$, the probe captures the error growth that occurs immediately after
an action perturbation; at $K{=}24$, it also captures the subsequent period in
which the task geometry limits further separation. The former therefore
characterizes the short-horizon cost of committing to a chunk, whereas the
latter asks whether that amplification persists over a longer execution
interval. This also raises a subtle consideration for shorter-chunk
execution. If the initial amplification is repeatedly followed by another
chunk boundary before the divergence has reached its saturated regime, then
using multiple short chunks does not necessarily allow the system to benefit
from the subsequent stabilization observed at longer horizons. In particular,
shortening a chunk without correspondingly increasing the frequency of
corrective replanning could repeatedly expose the system to the high-growth
portion of the trajectory, potentially inducing more error rather than
allowing an initially growing deviation to stabilize. Our measurements do not
establish this accumulation effect directly, since each probe introduces a
single perturbation and measures its subsequent evolution; however, they show
why the relationship between chunk length and error propagation cannot be
inferred from the short-window growth rate alone. For this reason, claims
about the fraction of unstable states must always be stated together with the
fitting window, and we use $K{=}24$ as the primary window while reporting the
matched $K{=}8$ analysis to ensure a fair comparison across stamp types.

The result also clarifies how the contraction-based lower bound of
\citet{zhang2025chunking} should be interpreted. Their guarantee requires
\begin{equation}
  k \;\ge\; k^{\star} \;=\;
  \log(1/\rho)^{-1}\,\log\!\big(\mathrm{poly}(L, C_{\mathrm{ISS}})\big),
  \label{eq:kstar}
\end{equation}
where the contraction rate $\rho$ determines the required chunk length.
Our measurements provide an empirical qualification of this picture: errors
can amplify during the initial portion of execution without continuing to
amplify over the full chunk horizon. Thus, whether a state appears unstable
depends not only on the state itself but also on how long the system is
observed after the perturbation. A short chunk can terminate during the
initial amplification phase, whereas a longer execution may include the
subsequent saturation of the divergence. We therefore do not interpret the
window result as establishing an optimal chunk length or as validating the
bound. In particular, our probe introduces one perturbation and measures its
subsequent evolution, so it does not determine whether errors introduced at
successive replans accumulate across chunks.

Read against execution, this has a simple implication for chunk length. A
short open-loop chunk may terminate while the error is still growing, whereas a
longer chunk continues through the subsequent plateau and thus amortises the
same initial transient over more executed steps. Consequently, when considering
the error incurred by a single chunk, a shorter chunk can actually be less
favourable than a longer one.

\subsubsection{The two axes together}
\label{sec:cl-3x3-para}

The joint labels reveal whether replanning changes the propagation of an injected error that would otherwise grow during open-loop execution. Across all three benchmark families, the $[\text{OL unstable, CL stable}]$ cell is near zero, indicating that states with error amplification under open-loop execution rarely show contraction when the policy is allowed to replan. Instead, these states are more often classified as deadband or remain unstable under closed-loop execution. Thus, within the measured propagation horizon, replanning does not generally convert error growth into error contraction.

The two labels therefore characterize complementary modes of error propagation: open-loop stability measures how the dynamics propagate an injected error without replanning, while closed-loop stability measures how the policy--dynamics pair propagates the same perturbation when replanning is available. The results show that neither regime can be assumed to attenuate a small error. This finding concerns error propagation only; it does not by itself establish which execution regime yields better task performance or determine an optimal chunk length.

\begin{figure}[t]
\centering
\input{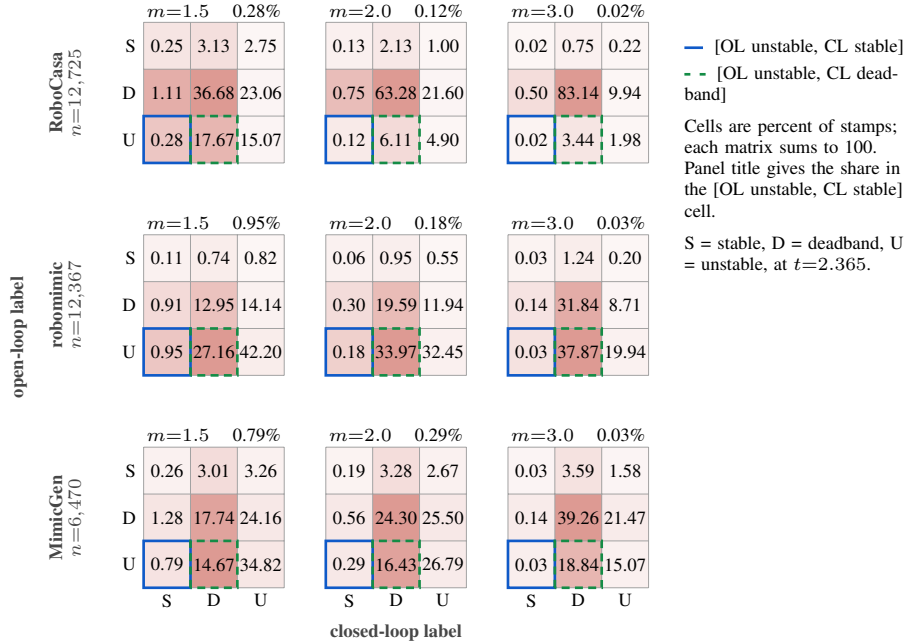}
\caption{Open-loop $\times$ closed-loop label matrices pooled by platform for $m=1.5,2.0,3.0$. Cells show the percentage of stamps in each joint label pair (S = stable, D = deadband, U = unstable). Blue outlines mark [OL unstable, CL stable] cells; green outlines mark [OL unstable, CL deadband] cells.}
\label{fig:3x3-k24}
\end{figure}

\subsection{Predicting the regime from observation}
\label{sec:results-pred}

We next ask whether the measured stability regime can be inferred from information available at a state. We train predictors separately on the three open-loop stamp types---\emph{demonstration}, \emph{rollout}, and \emph{policy} stamps---since they correspond to different state distributions and action sources.  Deadband stamps remain available as regression targets, but are excluded from classification because their rate is either too small to matter or insufficiently resolved from zero. Thus, regression is evaluated on all stamps, while classification is trained and evaluated only on stamps that clear both gates. Train and validation sets are split by demonstration rather than by individual stamp to avoid placing near-duplicate stamps from the same episode on opposite sides of the split. Each cell is trained with three random initialisations, and the reported values are averaged across seeds.

We consider two input settings. \emph{P} receives camera observations and proprioception over the observation window, asking whether the stability regime is reflected in the currently observed configuration. \emph{A1} additionally receives the action sequence being evaluated, providing information about the particular execution for which the propagation rate is measured. The same predictor architecture and training procedure are used in both settings; only the available inputs differ.

\subsubsection{Predicting open-loop stability}

Open-loop stability exhibits measurable state-dependent structure that can be
partially recovered from observation. Across the demonstration and rollout
stamps, the camera-and-proprioception predictor achieves AUROC values between
$0.63$ and $0.94$ on cells with sufficient negative-class support, with the
strongest results on \texttt{tool\_hang} and \texttt{mg-coffee}. Thus, the
observed configuration contains information about whether a small action error
is likely to amplify or decay, even without access to the simulator or the
actions used in the probe.

The regression is weaker than the classification. Measured against a constant
predictor that outputs the training-set median for every stamp, the head
recovers between a tenth and a quarter of the error, and on \texttt{can} it does
not improve on that baseline at all. In other words, the model is better at identifying which states are more prone to error amplification than at predicting how much the error will amplify. This suggests that observations contain enough information to distinguish stability regimes, but not enough to accurately reconstruct the propagation rate itself.

Adding the action sequence generally improves the prediction of $\lambda$, particularly in terms of regression error and rank correlation. This is expected because the measured open-loop rate depends on the particular sequence of actions being replayed. Providing that sequence therefore gives the predictor information about the execution that is not available from camera observations and proprioception alone. The improvement is not uniform across classification metrics, so we view A1 as providing additional information about the measured execution rather than as uniformly improving regime classification.

\begin{table}[t]
\centering
\small
\setlength{\tabcolsep}{5pt}
\begin{tabular}{llrrr|rrr}
\toprule
& & \multicolumn{3}{c|}{\textbf{Open-loop}} &
   \multicolumn{3}{c}{\textbf{Closed-loop}} \\
\cmidrule(lr){3-5}\cmidrule(lr){6-8}
Dataset & Input & MAE$_0$ & MAE & $\rho$ & MAE$_0$ & MAE & $\rho$ \\
\midrule
lift & P  & .0792 & \textbf{.0535} & .757 & .0322 & \textbf{.0178} & .819 \\
     & A1 & .0792 & \textbf{.0400} & .795 & .0322 & \textbf{.0164} & .841 \\
can  & P  & .0981 & \textbf{.0635} & .706 & .0395 & \textbf{.0323} & .520 \\
     & A1 & .0981 & \textbf{.0568} & .762 & .0395 & \textbf{.0305} & .553 \\
square & P  & .0877 & \textbf{.0790} & .510 & .0370 & \textbf{.0364} & .332 \\
       & A1 & .0877 & \textbf{.0742} & .560 & .0370 & \textbf{.0357} & .386 \\
tool\_hang & P  & .0887 & \textbf{.0687} & .627 & .0377 & .0393 & .239 \\
           & A1 & .0887 & \textbf{.0664} & .637 & .0377 & \textbf{.0347} & .333 \\
mg-coffee & P  & .0833 & .0842 & .149 & .0400 & \textbf{.0311} & .436 \\
          & A1 & .0833 & .0835 & .215 & .0400 & \textbf{.0317} & .455 \\
mg-nut & P  & .0959 & .0941 & .284 & .0391 & \textbf{.0373} & .374 \\
       & A1 & .0959 & .0927 & .324 & .0391 & \textbf{.0344} & .474 \\
lib-K3$^\dagger$ & P  & .0820 & \textbf{.0634} & .618 & -- & -- & -- \\
                 & A1 & .0820 & \textbf{.0609} & .648 & -- & -- & -- \\
\bottomrule
\end{tabular}
\caption{Prediction of open- and closed-loop propagation rates at the same
policy-replan stamps. Both heads use the same trunk and inputs; only the
target differs. $\mathrm{MAE}_0$ is the constant-median baseline, and
$\rho$ is Spearman correlation with the measured rate. Bold denotes
improvement over the constant-median baseline.}
\label{tab:pred-dual}
\end{table}

\subsubsection{Predicting closed-loop stability}

The closed-loop rate depends on both the current state and the behaviour of the deployed policy after the perturbation. Nevertheless, because the labels are generated under the same policy used at deployment, the rate may still contain predictable structure tied to the observed configuration. We therefore treat closed-loop prediction as a controlled target ablation: the predictor architecture, inputs, and policy-replan stamps are unchanged, while the target is switched from the open-loop to the closed-loop propagation rate.

The closed-loop rate is partially recoverable from these inputs. Across the evaluated tasks and input variants, the predictor improves over the constant-median baseline on most regression cells, while AUROC exceeds \(0.70\) in half of the cells for which both classes are present. The strongest results occur on lift and mg-coffee, whereas performance is weaker and closer to chance on several of the remaining tasks. The classification results should be interpreted cautiously because the stable closed-loop class is very small at \(m=2\), with only a few to tens of stable stamps per task and several single-class validation splits. Thus, the result is not that closed-loop stability is purely unpredictable; rather, some state-dependent structure is recoverable, but substantially less uniformly than for open-loop propagation. This is consistent with closed-loop stability depending not only on the configuration but also on the policy's subsequent behaviour.

\section{Limitations}
These results do not determine the best execution horizon by themselves:
stability is one factor alongside task performance, temporal consistency,
replanning cost, and other effects of action chunking. They instead provide a
state-dependent measure of the error-propagation consequences of continuing a
chunk versus replanning. If closed-loop reactivity is desired, our results
suggest that it should be trained explicitly, using perturbation- and
tree-coverage-oriented methods to expose policies to deviations they must
recover from, rather than expected to emerge reliably from standard imitation
learning.

\section{Conclusion}
\label{sec:conclusion}
We studied what happens after an action error enters a chunked imitation-learning
system, rather than how often the policy introduces one. By measuring error
propagation from individual states under both open-loop execution and
closed-loop replanning, we find that confidently stable states are uncommon,
while amplification is common among states for which the propagation rate is
resolved. Replanning frequently removes confident amplification, but rarely
produces confident contraction, suggesting that feedback does not generally
turn an injected deviation into a reliably shrinking one.

The measurements also show that error propagation is strongly state- and
horizon-dependent. Shorter fitting windows emphasize the initial amplification
of a perturbation, while longer windows include subsequent saturation, making
the measured rate a finite-horizon quantity rather than a single global
stability constant. More importantly, even within a single task, propagation
rates vary substantially across states, so the assumption of a single global
contraction rate is not supported empirically even at the task level, let alone
across an entire benchmark dataset.

Taken together, these results suggest that closed-loop error recovery should
not be expected to emerge reliably from standard imitation learning. If such
recovery is desired, it should be trained explicitly by exposing the policy to
perturbations and recovery trajectories through perturbation- and
tree-coverage-oriented training.

\bibliography{example}
\bibliographystyle{iclr2027_conference}

\appendix
\part*{Appendix}
\addcontentsline{toc}{part}{Appendix}
\section{Robustness of the Stability Measurements}
\label{app:robustness}

The main text uses $m{=}2.0$ and $K{=}24$ as the default settings. This
appendix examines how the stability labels change when the amplification
threshold is varied. Unless otherwise stated, all analyses reuse the stored
per-branch divergence trajectories and require no additional simulation.

\begin{figure}[t]
\centering
\input{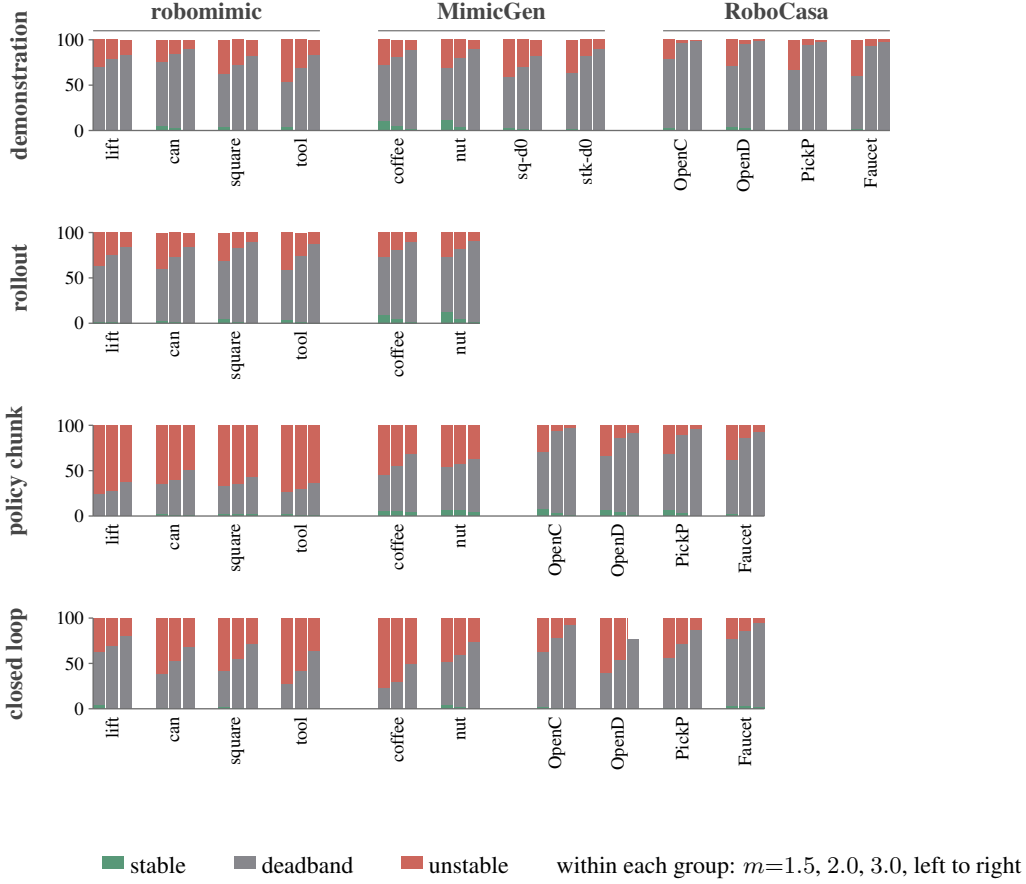}
\caption{Task-level breakdown of stability labels across the four stamp types.
For each task, the three adjacent bars correspond to Gate-1 thresholds
$m=1.5$, $2.0$, and $3.0$; each bar is partitioned into stable,
deadband, and unstable stamps.}
\label{fig:ol-stack}
\end{figure}

\subsection{Sensitivity to the amplification threshold}
\label{app:shares}

We repeat the three-way labeling at
$m\in{1.5,2.0,3.0}$. Increasing $m$ raises the Gate~1 threshold
$\delta_m=\log(m)/K$, moving states with rates close to zero into the
deadband.

Table~\ref{tab:app-ol-shares} reports the resulting open-loop label shares.
The fractions change with $m$, but the overall pattern is stable: confidently
stable states remain uncommon, while unstable states remain prevalent among
stamps whose rates are sufficiently far from zero.


\begin{table}[t]
\centering
\small
\setlength{\tabcolsep}{4.5pt}
\begin{tabular}{llrrrrrrrrr}
\toprule
& & \multicolumn{3}{c}{$m{=}1.5$}
& \multicolumn{3}{c}{$m{=}2.0$ (primary)}
& \multicolumn{3}{c}{$m{=}3.0$} \\
\cmidrule(lr){3-5}
\cmidrule(lr){6-8}
\cmidrule(lr){9-11}
Dataset & Stamps & S & D & U & S & D & U & S & D & U \\
\midrule
\multicolumn{11}{l}{\emph{robomimic}} \\
lift & demos
& 0.2 & 70.0 & 29.8
& 0.0 & 78.9 & 21.1
& 0.0 & 84.1 & 15.9 \\
lift & rollouts
& 2.4 & 60.8 & 36.8
& 1.4 & 74.1 & 24.5
& 0.5 & 83.4 & 16.1 \\
lift & policy
& 0.2 & 24.0 & 75.9
& 0.2 & 28.1 & 71.7
& 0.2 & 37.9 & 61.9 \\
can & demos
& 4.7 & 71.4 & 23.9
& 2.9 & 82.0 & 15.1
& 1.1 & 89.2 & 9.7 \\
can & rollouts
& 3.0 & 57.4 & 39.6
& 1.7 & 71.3 & 27.0
& 0.7 & 83.9 & 15.4 \\
can & policy
& 2.1 & 33.0 & 64.9
& 1.7 & 38.1 & 60.2
& 1.5 & 49.1 & 49.4 \\
square & demos
& 3.9 & 59.0 & 37.2
& 1.5 & 70.6 & 27.9
& 0.7 & 82.2 & 17.1 \\
square & rollouts
& 4.6 & 64.6 & 30.9
& 2.0 & 80.6 & 17.4
& 0.8 & 89.1 & 10.2 \\
square & policy
& 2.4 & 30.5 & 67.1
& 2.4 & 33.5 & 64.1
& 2.2 & 41.4 & 56.3 \\
tool\_hang & demos
& 4.1 & 49.4 & 46.6
& 1.4 & 67.4 & 31.3
& 0.2 & 83.5 & 16.3 \\
tool\_hang & rollouts
& 4.0 & 55.2 & 40.8
& 1.4 & 73.3 & 25.3
& 0.2 & 87.5 & 12.3 \\
tool\_hang & policy
& 2.0 & 24.9 & 73.1
& 1.9 & 27.9 & 70.1
& 1.9 & 34.8 & 63.4 \\
\midrule
\multicolumn{11}{l}{\emph{MimicGen}} \\
coffee & demos
& 10.5 & 61.7 & 27.8
& 5.5 & 76.1 & 18.4
& 2.1 & 87.2 & 10.7 \\
coffee & rollouts
& 9.2 & 63.7 & 27.1
& 4.9 & 76.2 & 18.9
& 1.6 & 87.8 & 10.6 \\
coffee & policy
& 6.1 & 38.9 & 55.0
& 5.8 & 49.4 & 44.8
& 5.2 & 63.3 & 31.5 \\
nut\_assembly & demos
& 11.5 & 57.6 & 30.9
& 4.7 & 75.2 & 20.2
& 1.3 & 88.9 & 9.8 \\
nut\_assembly & rollouts
& 12.5 & 60.3 & 27.2
& 5.4 & 76.7 & 17.9
& 1.6 & 88.9 & 9.5 \\
nut\_assembly & policy
& 7.0 & 47.4 & 45.6
& 6.5 & 51.3 & 42.2
& 5.1 & 58.5 & 36.4 \\
square\_d0 & demos
& 3.5 & 55.6 & 40.9
& 1.6 & 68.6 & 29.8
& 0.4 & 82.3 & 17.3 \\
stack\_d0 & demos
& 1.4 & 62.7 & 35.9
& 0.3 & 81.7 & 18.0
& 0.0 & 90.0 & 10.0 \\
\midrule
\multicolumn{11}{l}{\emph{RoboCasa}} \\
OpenCabinet & demos
& 2.9 & 76.1 & 21.0
& 1.2 & 95.6 & 3.2
& 0.3 & 98.8 & 0.9 \\
OpenCabinet & policy
& 8.0 & 62.8 & 29.2
& 4.1 & 89.3 & 6.6
& 1.0 & 96.3 & 2.8 \\
OpenDrawer & demos
& 4.5 & 66.7 & 28.7
& 2.5 & 93.2 & 4.3
& 0.9 & 97.7 & 1.4 \\
OpenDrawer & policy
& 7.0 & 59.8 & 33.3
& 4.6 & 81.9 & 13.5
& 1.9 & 89.9 & 8.1 \\
PickPlaceToCab. & demos
& 0.5 & 67.0 & 32.5
& 0.1 & 94.0 & 5.9
& 0.0 & 98.3 & 1.7 \\
PickPlaceToCab. & policy
& 6.9 & 61.5 & 31.6
& 3.5 & 85.9 & 10.6
& 0.6 & 95.3 & 4.1 \\
TurnOnSinkFaucet & demos
& 2.0 & 58.9 & 39.1
& 1.0 & 92.3 & 6.7
& 0.3 & 97.3 & 2.4 \\
TurnOnSinkFaucet & policy
& 2.7 & 59.4 & 38.0
& 0.8 & 85.3 & 13.8
& 0.4 & 92.9 & 6.7 \\
\bottomrule
\end{tabular}
\caption{Open-loop label shares in percent at $t{=}2.365$, across all three
amplification settings. S = stable, D = deadband, U = unstable; rows sum to
100. Each row is evaluated at its stored fitting window: $K{=}24$ for
demonstration and rollout stamps, and the emitted chunk length for policy
stamps.}
\label{tab:app-ol-shares}
\end{table}

\subsection{Full open-loop rate distributions}
\label{app:quantiles}

Table~\ref{tab:app-ol-quantiles} reports the median, interquartile range,
and $5$--$95$ percentile range of $\lambda_{\mathrm{mean}}$ for each dataset
and stamp type. The medians are positive across the open-loop conditions,
but their magnitudes vary substantially across tasks, and most distributions
contain both positive and negative rates. The standard-error column reports
the typical uncertainty of the estimated rate.

\begin{table}[t]
\centering
\small
\setlength{\tabcolsep}{4pt}
\begin{tabular}{llrrlrrrr}
\toprule
Dataset & Stamps & $K$ & median & IQR & p5 & p95 &
$\lambda_{\mathrm{se}}$ & $e^{16\lambda}_{\mathrm{med}}$ \\
\midrule
\multicolumn{9}{l}{\emph{robomimic}} \\
lift & demos & 24 & .015 & .005--.035 & $-$.012 & .184 & .0076 & 1.26 \\
lift & rollouts & 24 & .019 & .005--.044 & $-$.020 & .127 & .0084 & 1.36 \\
lift & policy & 8 & .090 & .059--.139 & .021 & .458 & .0216 & 4.23 \\
can & demos & 24 & .009 & $-$.003--.028 & $-$.032 & .115 & .0059 & 1.15 \\
can & rollouts & 24 & .019 & .002--.048 & $-$.022 & .116 & .0078 & 1.36 \\
can & policy & 8 & .073 & .032--.132 & $-$.016 & .536 & .0172 & 3.22 \\
square & demos & 24 & .017 & .001--.052 & $-$.026 & .128 & .0082 & 1.32 \\
square & rollouts & 24 & .018 & .004--.032 & $-$.029 & .104 & .0067 & 1.32 \\
square & policy & 8 & .091 & .046--.157 & $-$.050 & .385 & .0253 & 4.30 \\
tool\_hang & demos & 24 & .026 & .003--.053 & $-$.025 & .114 & .0084 & 1.50 \\
tool\_hang & rollouts & 24 & .021 & .003--.045 & $-$.024 & .101 & .0077 & 1.39 \\
tool\_hang & policy & 8 & .113 & .048--.180 & $-$.015 & .348 & .0216 & 6.13 \\
\midrule
\multicolumn{9}{l}{\emph{MimicGen}} \\
coffee & demos & 24 & .007 & $-$.008--.033 & $-$.047 & .103 & .0072 & 1.13 \\
coffee & rollouts & 24 & .009 & $-$.003--.032 & $-$.043 & .105 & .0069 & 1.15 \\
coffee & policy & 8 & .045 & .018--.094 & $-$.134 & .301 & .0191 & 2.06 \\
nut\_assembly & demos & 24 & .009 & $-$.011--.038 & $-$.043 & .097 & .0076 & 1.15 \\
nut\_assembly & rollouts & 24 & .006 & $-$.012--.033 & $-$.046 & .099 & .0078 & 1.11 \\
nut\_assembly & policy & 8 & .046 & .000--.116 & $-$.121 & .329 & .0230 & 2.08 \\
square\_d0 & demos & 24 & .020 & .001--.054 & $-$.024 & .122 & .0083 & 1.37 \\
stack\_d0 & demos & 24 & .019 & .006--.035 & $-$.014 & .115 & .0069 & 1.35 \\
\midrule
\multicolumn{9}{l}{\emph{RoboCasa}} \\
OpenCabinet & demos & 24 & .016 & .004--.024 & $-$.022 & .038 & .0062 & 1.29 \\
OpenCabinet & policy & 24 & .019 & .005--.028 & $-$.040 & .051 & .0060 & 1.35 \\
OpenDrawer & demos & 24 & .018 & .004--.027 & $-$.028 & .042 & .0056 & 1.34 \\
OpenDrawer & policy & 24 & .020 & .003--.031 & $-$.043 & .089 & .0069 & 1.37 \\
PickPlaceToCab. & demos & 24 & .020 & .012--.028 & $-$.006 & .046 & .0048 & 1.39 \\
PickPlaceToCab. & policy & 24 & .019 & .006--.029 & $-$.035 & .064 & .0054 & 1.36 \\
TurnOnSinkFaucet & demos & 24 & .023 & .014--.030 & $-$.013 & .050 & .0047 & 1.44 \\
TurnOnSinkFaucet & policy & 24 & .022 & .013--.033 & $-$.021 & .087 & .0061 & 1.43 \\
\bottomrule
\end{tabular}
\caption{Full $\lambda_{\mathrm{mean}}$ quantiles on the open-loop axis.
The table extends the main-text rate summaries with the interquartile and
$5$--$95$ percentile ranges. $e^{16\lambda}_{\mathrm{med}}$ converts the
median rate to error amplification over 16 steps. For rows measured at
$K{=}8$, this extrapolates beyond the measured window and should therefore be
read only as a derived quantity, not as an observed 16-step trajectory.}
\label{tab:app-ol-quantiles}
\end{table}

\subsection{Sensitivity to the fitting horizon}
\label{app:k16}

The main text compares $K{=}8$ and $K{=}24$. We use $K{=}16$ as an intermediate
window to determine whether the observed change is specific to either endpoint.
All quantities below are refit from the same stored per-branch log-distance
matrices, so no new simulation is involved.





\subsection{The temporal origin of the horizon effect}
\label{app:window-rate}

\begin{figure}[t]
\centering
\begin{tikzpicture}[x=1cm,y=1cm,font=\tiny]
\draw[cline!50] (0,0) rectangle (7.0,2.2);
\draw[cline!50] (0.437,0) -- (0.437,-0.06) node[below,inner sep=1pt] {-0.1};
\draw[cline!50] (1.531,0) -- (1.531,-0.06) node[below,inner sep=1pt] {-0.05};
\draw[cline!50] (2.625,0) -- (2.625,-0.06) node[below,inner sep=1pt] {0};
\draw[cline!50] (3.719,0) -- (3.719,-0.06) node[below,inner sep=1pt] {0.05};
\draw[cline!50] (4.812,0) -- (4.812,-0.06) node[below,inner sep=1pt] {0.1};
\draw[cline!50] (5.906,0) -- (5.906,-0.06) node[below,inner sep=1pt] {0.15};
\draw[cline!50] (7.000,0) -- (7.000,-0.06) node[below,inner sep=1pt] {0.2};
\draw[black,very thick] (3.573,0) -- (3.573,2.2);
\draw[black,thin] (1.677,0) -- (1.677,2.2);
\draw[stab,thick] (0.000,0.000) -- (0.109,0.000) -- (0.109,0.000) -- (0.219,0.000) -- (0.219,0.001) -- (0.328,0.001) -- (0.328,0.002) -- (0.438,0.002) -- (0.438,0.001) -- (0.547,0.001) -- (0.547,0.003) -- (0.656,0.003) -- (0.656,0.005) -- (0.766,0.005) -- (0.766,0.005) -- (0.875,0.005) -- (0.875,0.010) -- (0.984,0.010) -- (0.984,0.006) -- (1.094,0.006) -- (1.094,0.014) -- (1.203,0.014) -- (1.203,0.018) -- (1.312,0.018) -- (1.312,0.023) -- (1.422,0.023) -- (1.422,0.033) -- (1.531,0.033) -- (1.531,0.036) -- (1.641,0.036) -- (1.641,0.060) -- (1.750,0.060) -- (1.750,0.078) -- (1.859,0.078) -- (1.859,0.101) -- (1.969,0.101) -- (1.969,0.150) -- (2.078,0.150) -- (2.078,0.212) -- (2.188,0.212) -- (2.188,0.298) -- (2.297,0.298) -- (2.297,0.390) -- (2.406,0.390) -- (2.406,0.536) -- (2.516,0.536) -- (2.516,0.673) -- (2.625,0.673) -- (2.625,0.850) -- (2.734,0.850) -- (2.734,1.169) -- (2.844,1.169) -- (2.844,1.628) -- (2.953,1.628) -- (2.953,2.037) -- (3.062,2.037) -- (3.062,1.974) -- (3.172,1.974) -- (3.172,1.394) -- (3.281,1.394) -- (3.281,0.789) -- (3.391,0.789) -- (3.391,0.344) -- (3.500,0.344) -- (3.500,0.181) -- (3.609,0.181) -- (3.609,0.103) -- (3.719,0.103) -- (3.719,0.075) -- (3.828,0.075) -- (3.828,0.045) -- (3.938,0.045) -- (3.938,0.038) -- (4.047,0.038) -- (4.047,0.029) -- (4.156,0.029) -- (4.156,0.026) -- (4.266,0.026) -- (4.266,0.017) -- (4.375,0.017) -- (4.375,0.014) -- (4.484,0.014) -- (4.484,0.012) -- (4.594,0.012) -- (4.594,0.009) -- (4.703,0.009) -- (4.703,0.010) -- (4.812,0.010) -- (4.812,0.006) -- (4.922,0.006) -- (4.922,0.006) -- (5.031,0.006) -- (5.031,0.004) -- (5.141,0.004) -- (5.141,0.001) -- (5.250,0.001) -- (5.250,0.001) -- (5.359,0.001) -- (5.359,0.001) -- (5.469,0.001) -- (5.469,0.001) -- (5.578,0.001) -- (5.578,0.001) -- (5.688,0.001) -- (5.688,0.002) -- (5.797,0.002) -- (5.797,0.001) -- (5.906,0.001) -- (5.906,0.000) -- (6.016,0.000) -- (6.016,0.000) -- (6.125,0.000) -- (6.125,0.000) -- (6.234,0.000) -- (6.234,0.000) -- (6.344,0.000) -- (6.344,0.000) -- (6.453,0.000) -- (6.453,0.000) -- (6.562,0.000) -- (6.562,0.000) -- (6.672,0.000) -- (6.672,0.000) -- (6.781,0.000) -- (6.781,0.000) -- (6.891,0.000) -- (6.891,0.000) -- (7.000,0.000);
\draw[cline,thick] (0.000,0.036) -- (0.109,0.036) -- (0.109,0.006) -- (0.219,0.006) -- (0.219,0.008) -- (0.328,0.008) -- (0.328,0.007) -- (0.438,0.007) -- (0.438,0.014) -- (0.547,0.014) -- (0.547,0.009) -- (0.656,0.009) -- (0.656,0.014) -- (0.766,0.014) -- (0.766,0.019) -- (0.875,0.019) -- (0.875,0.027) -- (0.984,0.027) -- (0.984,0.024) -- (1.094,0.024) -- (1.094,0.029) -- (1.203,0.029) -- (1.203,0.034) -- (1.312,0.034) -- (1.312,0.043) -- (1.422,0.043) -- (1.422,0.051) -- (1.531,0.051) -- (1.531,0.071) -- (1.641,0.071) -- (1.641,0.084) -- (1.750,0.084) -- (1.750,0.082) -- (1.859,0.082) -- (1.859,0.141) -- (1.969,0.141) -- (1.969,0.157) -- (2.078,0.157) -- (2.078,0.170) -- (2.188,0.170) -- (2.188,0.223) -- (2.297,0.223) -- (2.297,0.257) -- (2.406,0.257) -- (2.406,0.308) -- (2.516,0.308) -- (2.516,0.355) -- (2.625,0.355) -- (2.625,0.407) -- (2.734,0.407) -- (2.734,0.473) -- (2.844,0.473) -- (2.844,0.550) -- (2.953,0.550) -- (2.953,0.634) -- (3.062,0.634) -- (3.062,0.732) -- (3.172,0.732) -- (3.172,0.952) -- (3.281,0.952) -- (3.281,1.088) -- (3.391,1.088) -- (3.391,1.201) -- (3.500,1.201) -- (3.500,1.244) -- (3.609,1.244) -- (3.609,1.083) -- (3.719,1.083) -- (3.719,0.848) -- (3.828,0.848) -- (3.828,0.659) -- (3.938,0.659) -- (3.938,0.447) -- (4.047,0.447) -- (4.047,0.294) -- (4.156,0.294) -- (4.156,0.198) -- (4.266,0.198) -- (4.266,0.107) -- (4.375,0.107) -- (4.375,0.075) -- (4.484,0.075) -- (4.484,0.048) -- (4.594,0.048) -- (4.594,0.044) -- (4.703,0.044) -- (4.703,0.029) -- (4.812,0.029) -- (4.812,0.022) -- (4.922,0.022) -- (4.922,0.020) -- (5.031,0.020) -- (5.031,0.008) -- (5.141,0.008) -- (5.141,0.010) -- (5.250,0.010) -- (5.250,0.012) -- (5.359,0.012) -- (5.359,0.010) -- (5.469,0.010) -- (5.469,0.011) -- (5.578,0.011) -- (5.578,0.007) -- (5.688,0.007) -- (5.688,0.008) -- (5.797,0.008) -- (5.797,0.006) -- (5.906,0.006) -- (5.906,0.005) -- (6.016,0.005) -- (6.016,0.003) -- (6.125,0.003) -- (6.125,0.003) -- (6.234,0.003) -- (6.234,0.003) -- (6.344,0.003) -- (6.344,0.001) -- (6.453,0.001) -- (6.453,0.002) -- (6.562,0.002) -- (6.562,0.001) -- (6.672,0.001) -- (6.672,0.000) -- (6.781,0.000) -- (6.781,0.001) -- (6.891,0.001) -- (6.891,0.003) -- (7.000,0.003);
\draw[unst,thick] (0.000,0.199) -- (0.109,0.199) -- (0.109,0.014) -- (0.219,0.014) -- (0.219,0.016) -- (0.328,0.016) -- (0.328,0.012) -- (0.438,0.012) -- (0.438,0.018) -- (0.547,0.018) -- (0.547,0.015) -- (0.656,0.015) -- (0.656,0.016) -- (0.766,0.016) -- (0.766,0.024) -- (0.875,0.024) -- (0.875,0.025) -- (0.984,0.025) -- (0.984,0.019) -- (1.094,0.019) -- (1.094,0.025) -- (1.203,0.025) -- (1.203,0.025) -- (1.312,0.025) -- (1.312,0.038) -- (1.422,0.038) -- (1.422,0.044) -- (1.531,0.044) -- (1.531,0.041) -- (1.641,0.041) -- (1.641,0.044) -- (1.750,0.044) -- (1.750,0.047) -- (1.859,0.047) -- (1.859,0.058) -- (1.969,0.058) -- (1.969,0.059) -- (2.078,0.059) -- (2.078,0.081) -- (2.188,0.081) -- (2.188,0.091) -- (2.297,0.091) -- (2.297,0.085) -- (2.406,0.085) -- (2.406,0.086) -- (2.516,0.086) -- (2.516,0.094) -- (2.625,0.094) -- (2.625,0.113) -- (2.734,0.113) -- (2.734,0.118) -- (2.844,0.118) -- (2.844,0.130) -- (2.953,0.130) -- (2.953,0.136) -- (3.062,0.136) -- (3.062,0.139) -- (3.172,0.139) -- (3.172,0.151) -- (3.281,0.151) -- (3.281,0.143) -- (3.391,0.143) -- (3.391,0.158) -- (3.500,0.158) -- (3.500,0.171) -- (3.609,0.171) -- (3.609,0.171) -- (3.719,0.171) -- (3.719,0.187) -- (3.828,0.187) -- (3.828,0.199) -- (3.938,0.199) -- (3.938,0.219) -- (4.047,0.219) -- (4.047,0.250) -- (4.156,0.250) -- (4.156,0.251) -- (4.266,0.251) -- (4.266,0.258) -- (4.375,0.258) -- (4.375,0.309) -- (4.484,0.309) -- (4.484,0.306) -- (4.594,0.306) -- (4.594,0.357) -- (4.703,0.357) -- (4.703,0.386) -- (4.812,0.386) -- (4.812,0.390) -- (4.922,0.390) -- (4.922,0.424) -- (5.031,0.424) -- (5.031,0.433) -- (5.141,0.433) -- (5.141,0.448) -- (5.250,0.448) -- (5.250,0.444) -- (5.359,0.444) -- (5.359,0.448) -- (5.469,0.448) -- (5.469,0.461) -- (5.578,0.461) -- (5.578,0.444) -- (5.688,0.444) -- (5.688,0.435) -- (5.797,0.435) -- (5.797,0.416) -- (5.906,0.416) -- (5.906,0.380) -- (6.016,0.380) -- (6.016,0.348) -- (6.125,0.348) -- (6.125,0.333) -- (6.234,0.333) -- (6.234,0.326) -- (6.344,0.326) -- (6.344,0.271) -- (6.453,0.271) -- (6.453,0.243) -- (6.562,0.243) -- (6.562,0.216) -- (6.672,0.216) -- (6.672,0.202) -- (6.781,0.202) -- (6.781,0.189) -- (6.891,0.189) -- (6.891,1.245) -- (7.000,1.245);
\node[below] at (3.50,-0.4) {$\lambda_{mean}$ (share per 0.005 bin; mass clipped to range)};
\node[right,font=\tiny] at (0.08,2.38) {\textcolor{unst}{--- $K{=}8$}\ \textcolor{cline}{--- $K{=}16$}\ \textcolor{stab}{--- $K{=}24$}\quad vertical: $\pm0.0433$ ($m{=}2.0$)};
\end{tikzpicture}
\caption{$\lambda_{\mathrm{mean}}$ on the open-loop axis at $K{=}8$, $16$, and
$24$ for the \texttt{OpenCabinet} demonstration stamps ($n{=}33{,}238$),
refit from the same stored per-branch log-distance matrices with no new
simulation. The three curves therefore describe identical stamps and differ
only in the length of the fitting interval. Verticals mark $\pm0.0433$, the
primary Gate-1 threshold at $m{=}2.0$.}
\label{fig:window-lam}
\end{figure}
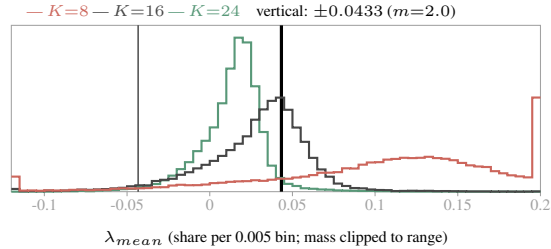

The dependence on $K$ is also visible directly in the fitted rate. Figure
\ref{fig:window-lam} overlays the rate distributions obtained at
$K{=}8$, $16$, and $24$ for the same OpenCabinet demonstration stamps. The
distributions shift toward larger positive rates as the fitting window
shortens, while the stable side remains sparse.

This behavior is consistent with the temporal profile of the divergence
trajectories. The initial portion of the trajectory typically contains the
strongest separation, while later steps show a weaker increase as the two
branches approach a task-dependent plateau. A short window therefore places
more weight on the initial growth, whereas a longer window averages that
growth with the later plateau and pulls the fitted rate toward zero.

The measured $\lambda$ should consequently be interpreted as a finite-horizon
summary rather than as a time-invariant property of a state. The same
perturbation trajectory can yield different numerical rates when evaluated
over different portions of its evolution.

\section{Predicting the Stability Regime}
\label{app:predictor}

\subsection{Architecture and training}
\label{app:predictor-training}

Both predictor variants use the same frozen V-JEPA~2 visual representation and
proprioceptive input described in Section~\ref{sec:predictors}. The P variant
uses observation features only, while A1 additionally receives the action chunk
associated with the state. The two variants therefore differ only in the
information available to the predictor.

Data are split by demonstration rather than by individual stamp, preventing
temporally adjacent states from the same episode from appearing in both the
training and validation sets. Each configuration is trained with three random
initializations, and reported results are averaged across seeds.

Deadband states provide a regression target but no unambiguous classification
target, so classification is evaluated only on stamps that clear both gates.
For the dual-axis experiment, both heads are trained on the same states, with
the open-loop head predicting the planned-chunk propagation rate and the
closed-loop head predicting the closed-loop propagation rate.

\subsection{Open-loop prediction}
\label{app:pred-ol}

Table~\ref{tab:pred-ol-app} gives the complete open-loop prediction results.
The classifier generally ranks states substantially better than a constant
baseline, particularly for the larger demonstration datasets. The strongest
results occur on \texttt{tool\_hang}, \texttt{mg-coffee}, and the RoboCasa
demonstration cells, while smaller datasets show more variation.

The regression target is harder to recover precisely. The predicted rates
improve on the constant-median baseline in many cells, but the improvement is
smaller than the ranking performance. This distinction is important for the
intended use of the predictor: the observations contain enough information to
identify states with different stability regimes, but not necessarily enough
to estimate a precise numerical $\lambda$ from a single observation.

Several AUROC values are based on very small negative classes. These cases
should therefore be interpreted cautiously and as part of the broader
cross-dataset pattern rather than as independent evidence from a single split.

\begin{table}[t]
\centering
\small
\setlength{\tabcolsep}{4.5pt}
\begin{tabular}{llrrrrl}
\toprule
Dataset & var & base MAE & MAE & $\rho$ & AUROC & neg (per seed) \\
\midrule
\multicolumn{7}{l}{\emph{demonstration stamps}} \\
lift & P & .0351 & .0272 & .704 & -- & 0; 0; 0 \\
lift & A1 & .0351 & .0216 & .718 & -- & 0; 0; 0 \\
can & P & .0281 & .0284 & .392 & .722 & 50; 49; 46 \\
can & A1 & .0281 & .0280 & .426 & .797 & 50; 49; 46 \\
square & P & .0343 & .0309 & .526 & .712 & 40; 45; 45 \\
square & A1 & .0343 & .0289 & .593 & .830 & 40; 45; 45 \\
tool\_hang & P & .0325 & .0260 & .594 & .909 & 112; 148; 140 \\
tool\_hang & A1 & .0325 & .0247 & .651 & .942 & 112; 148; 140 \\
mg-coffee & P & .0311 & .0247 & .650 & .941 & 713; 723; 700 \\
mg-coffee & A1 & .0311 & .0226 & .690 & .947 & 713; 723; 700 \\
OpenCabinet & P & .0139 & .0130 & .358 & .810 & 104; 90; 56 \\
OpenCabinet & A1 & .0139 & .0124 & .443 & .909 & 104; 90; 56 \\
OpenDrawer & P & .0165 & .0150 & .367 & .800 & 117; 116; 125 \\
OpenDrawer & A1 & .0165 & .0140 & .483 & .894 & 117; 116; 125 \\
PickPlaceToCab. & P & .0114 & .0113 & .186 & .667 & 1; 5; 2 \\
PickPlaceToCab. & A1 & .0114 & .0106 & .389 & .846 & 1; 5; 2 \\
TurnOnSinkFaucet & P & .0132 & .0134 & .220 & .857 & 23; 42; 40 \\
TurnOnSinkFaucet & A1 & .0132 & .0127 & .336 & .864 & 23; 42; 40 \\
\midrule
\multicolumn{7}{l}{\emph{rollout stamps}} \\
lift & P & .0329 & .0309 & .404 & .631 & 21; 11; 4 \\
lift & A1 & .0329 & .0300 & .492 & .833 & 21; 11; 4 \\
can & P & .0305 & .0284 & .449 & .830 & 35; 38; 28 \\
can & A1 & .0305 & .0279 & .494 & .842 & 35; 38; 28 \\
\midrule
\multicolumn{7}{l}{\emph{policy stamps}} \\
OpenCabinet & P & .0196 & .0247 & .097 & .712 & 21; 17; 27 \\
OpenCabinet & A1 & .0196 & .0208 & .213 & .822 & 21; 17; 27 \\
OpenDrawer & P & .0242 & .0287 & .218 & .697 & 25; 24; 19 \\
OpenDrawer & A1 & .0242 & .0248 & .309 & .767 & 25; 24; 19 \\
PickPlaceToCab. & P & .0213 & .0258 & .206 & .706 & 29; 19; 13 \\
PickPlaceToCab. & A1 & .0213 & .0214 & .345 & .886 & 29; 19; 13 \\
TurnOnSinkFaucet & P & .0247 & .0246 & .190 & .626 & 4; 10; 13 \\
TurnOnSinkFaucet & A1 & .0247 & .0240 & .277 & .774 & 4; 10; 13 \\
\bottomrule
\end{tabular}
\caption{Open-loop prediction results, means over three seeds. \emph{base MAE}
is a constant-median predictor on the same validation split. $\rho$ is the
Spearman correlation between predicted and measured $\lambda$, and AUROC is
computed over the confident subset. \emph{neg} gives the number of negative
examples behind each seed's AUROC.}
\label{tab:pred-ol-app}
\end{table}

\end{document}